\documentclass[11pt]{article}
\usepackage{acl}

\usepackage{microtype}
\usepackage{array}
\usepackage{tabularx}
\usepackage{colortbl}
\usepackage{todonotes}
\usepackage{xspace}
\usepackage{times}
\usepackage{latexsym}
\usepackage{mathptmx}
\usepackage{multirow}
\usepackage{enumitem}

\usepackage{xcolor}
\usepackage{soul}
\sethlcolor{yellow}

\usepackage{bbold}

\usepackage[utf8]{inputenc}
\usepackage[LAE,T1]{fontenc}

\usepackage[arabic,english]{babel}
\usepackage{ragged2e}
\newcommand{\ab}{\selectlanguage{arabic}}
\newcommand{\en}{\selectlanguage{english}}

\usepackage{tikz}
\usepackage{amsmath}
\usetikzlibrary{shapes.geometric, arrows.meta, positioning, calc}
\usepackage{dblfloatfix}

\usepackage{changes}
\usepackage{cuted}

\title{MASRAD: Arabic Terminology Management Corpora with Semi-Automatic Construction}

\author{ Mahdi Nasser  \and  Laura Sayyah \and Fadi A. Zaraket \\
Arab Center for Research and Policy Studies, Doha \\
         \texttt{{\{mnasser,lsayyah,fzaraket\} }@dohainstitute.edu.qa}
        }

\begin{document}
\maketitle

\begin{abstract}

This paper presents MASRAD, a terminology dataset for Arabic terminology management, and a method with supporting tools for its semi-automatic construction. 
The entries in MASRAD are $(f,a)$ pairs of 
foreign (non-Arabic) terms $f$, appearing 
in  specialized, academic and field-specific books 
next to their Arabic $a$ counterparts. 
MASRAD-Ex systematically extracts these pairs as a first step to 
construct MASRAD. 
MASRAD helps improving {\em term consistency } in academic translations and  specialized Arabic documents, and automating cross-lingual text processing.
MASRAD-Ex leverages translated terms organically occurring in Arabic books, and considers several candidate pairs for each term phrase. 
The candidate Arabic terms occur next to the foreign terms, and vary in length. MASRAD-Ex computes lexicographic, phonetic, morphological, and semantic similarity metrics for each candidate pair, and uses heuristic, machine learning, and machine learning with post-processing approaches to decide on the best candidate. 
This paper presents MASRAD after thorough expert review and makes it 
available to the interested research community. 
The best performing MASRAD-Ex approach achieved 90.5\% precision and 92.4\% recall.

\end{abstract}

\section{Introduction}

{\em Terminology management} concerns
(i) specifying terms used upon the introduction of novel concepts to a language,
and
(ii) enforcing the use, hence
(iii) ensuring {\em consistency} and correctness.
Translation from foreign languages to Arabic sources significant  terminology inconsistency due to the diversity of translation
experts and the lack of reference parallel term corpora. Recent systems like TURJUMAN~\cite{nagoudi2022turjuman} have improved Arabic machine translation quality, but do not focus on terminology alignment or consistency.
Linguistic and structural complexities affect consistency and clarity of translations and may hinder access to knowledge.

The first and most important step is to build a termbase and provide it as a reference to Arabic scholars and experts. 

We present MASRAD, 
a large dataset of parallel terms with pairs of foreign  terms
and their corresponding Arabic counterparts, sourced into use in 
professional, academic and scholastic Arabic contexts. 
We make MASRAD available for the research community and
to authors looking for consistent Arabic terms when introducing 
a novel concept to Arabic. 

Recent advances in natural language processing (NLP) and computational linguistics (CL) have paved the way for automated systems that assist in cross-lingual text processing. 
However, existing systems, including LLMs, often struggle with the nuances of terminology in cross-lingual specific expert fields, where precision is paramount. 
This paper constructs MASRAD, and presents a novel methodology and
a supporting framework (MASRAD-Ex) to extract and curate 
parallel foreign-Arabic terms from existing Arabic books. 

When a foreign term occurs in Arabic scholarly writing, 
it usually occurs between parentheses, 
and it is usually preceded by its Arabic translation.
Two problems arise. 
(i) The exact scope of the Arabic term is ambiguous, 
and 
(ii) often times authors (even the same author) use Arabic terms inconsistently for the same non-Arabic ones. 
Editors and reviewers, 
who are usually expert researchers themselves, 
are required to manually solve these laborious and
time consuming tasks during lengthy editing iterations.

MASRAD and MASRAD-Ex help automate these two tasks to reduce review and edit efforts and help reviewers and editors concentrate on more important research and knowledge generation tasks. 

The problem is formulated as follows. Consider  a sequence  $s=\langle ( f ) w_1 w_2 w_3 \ldots w_n \rangle$ representing a sentence with a foreign term $f$ between parentheses, and preceded
from right to left with a sequence of Arabic words $w_1, \ldots w_n$.
We denote the candidate terms possible to match $f$ as the set of strings $S(f)=\{s_i: s_i=w_1 \ldots w_i, i\in[1,n]\}$ ending before term $f$. 
MASRAD-Ex extracts these candidates from the books, and 
identifies the target candidate term $s_t\in S(f) $ 
best matching $f$. 
MASRAD-Ex  extracts linguistic features for  each candidate match $(s_i,f)$. 
It  interpolates the features and aggregates their values into a resulting score, and then ranks the candidate terms. 

This work addresses the challenge of terminology mapping
and reduces  language barriers, facilitating a greater exchange of knowledge and ideas across languages and cultures.

We make MASRAD and the constructed dataset of feature vectors (used for training/testing) available online for the research community upon request\footnote{{https://github.com/mnasser-dru/MASRAD}}.
%
To the best of our knowledge, this approach to termbase building for Arabic is novel, and no comparable datasets currently exist.

\subsection{Motivating example}
Consider the following two contexts.
They contain the same underlined french term 
\textbf{l’ethnocentrisme} (The ethnocentrism) with different translations. 
\begin{itemize}[noitemsep,itemindent=0pt,leftmargin=1pt,topsep=0pt,parsep=0pt,partopsep=0pt]
    \item {
        Context 1:
        
        \ab ولا يتحول نقده الحاد للمركزية - الإثنية \foreignlanguage{english}{\underline{(l’ethnocentrisme)}} هنا إلى نسبية ثقافية، بل إلى كونية عقلية: <(...) نقبل بأنّ الأفكار والمبادئ العقلية فطرية لدينا>.
    
        \en
        
        \textit{Gloss: "His sharp criticism of ethnocentrism (l’ethnocentrisme) does not turn into cultural relativism, but rather into rational universality."}
    }
    \item {
        Context 2:
        
       \ab انتقد كثيراً النزعة الإثنية المركزية \foreignlanguage{english}{\underline{(l’ethnocentrisme)}} التي تريد <إحلال أفكارنا الأوروبية محل أفكار الإنسان البدائي عن العالم والمجتمع>، وتحذّر من خطر <نسبة الأفكار المتقدمة إلى الإنسان البدائي لأنها تتجاوز إمكاناته العقلية>.

       \en

        \textit{Gloss: "He strongly criticized the ethnocentric tendency (l’ethnocentrisme) that seeks to replace primitive human ideas about the world and society with our European ideas."}        
    }
\end{itemize}

MASRAD-Ex computers the following candidates per context. 
\textbf{Candidates for Context 1:}
\begin{itemize}
[noitemsep,topsep=0pt,parsep=0pt,partopsep=0pt]
    \item \textAR{ولا يتحول نقده الحاد للمركزية - الإثنية} 
    \textit{Gloss: "And his sharp criticism of ethnocentrism does not turn into..."}
    \item \textAR{يتحول نقده الحاد للمركزية - الإثنية} 
    \textit{Gloss: "His sharp criticism of ethnocentrism turns into..."}
    \item \textAR{نقده الحاد للمركزية - الإثنية}
    \textit{Gloss: "His sharp criticism of ethnocentrism"}
    \item  \textAR{الحاد للمركزية - الإثنية} 
    \textit{Gloss: "The sharp ethnocentrism"}
    \item  \textAR{للمركزية - الإثنية}  \hspace{0.2cm} (target term)  
    \textit{Gloss: "Ethnocentrism"}
    \item  \textAR{الإثنية} 
    \textit{Gloss: "Ethnicity"}
\end{itemize}

\noindent

\textbf{Candidates for Context 2:}
\begin{itemize}
[noitemsep,topsep=0pt,parsep=0pt,partopsep=0pt]
    \item \textAR{انتقد كثيراً النزعة الإثنية المركزية} 
    \textit{Gloss: "He strongly criticized the ethnocentric tendency"}
    \item \textAR{كثيراً النزعة الإثنية المركزية} 
    \textit{Gloss: "The strong ethnocentric tendency"}
    \item  \textAR{النزعة الإثنية المركزية}
    \textit{Gloss: "The ethnocentric tendency"}
    \item  \textAR{الإثنية المركزية}  \hspace{0.2cm} (target term)
    \textit{Gloss: "Ethnocentrism"}
    \item  \textAR{المركزية} 
    \textit{Gloss: "Centralism"}
\end{itemize}

The two contexts and the extracted translation candidate terms showcase the inconsistency across  contexts, and also expose potential adoption of different candidates by readers of each context. 
Thus amplifying the problem.

This paper makes the following contributions:
\begin{itemize}
    \item We propose a novel method for terminology mapping that leverages naturally occurring term translations, combining NLP and data driven  techniques.
    \item We build and provide MARSAD,  a unique annotated dataset for terminology extraction, which can serve as a valuable resource for the research community.
    \item We build and evaluate MASRAD-Ex including  three approaches: heuristic, machine learning (ML), and ML 
    with post-processing demonstrating the effectiveness of these methods in addressing terminology mapping challenges. 
\end{itemize}
The results are impressive and industry/deployment level,  convincing for deployment. The final model runs locally, requires no LLM interaction, and thus requires no data exposure or IP risks.


We discuss MASRAD in Section~\ref{s:d}, the methodology in Section~\ref{m}, present and discuss results in Section~\ref{s:results}, and  cover related work in ~\ref{s:rel}, 
We then discuss future directions and conclude. 

\section{Dataset creation}
\label{s:d}

\newcolumntype{C}{>{\centering\arraybackslash}m{3.5cm}}
\begin{table}[t]
    \caption{Statistics of all processed books}
    \label{t:books}
    \centering
    \small
\vspace{-1em}
    \begin{tabular}{|C|C|}
    \hline
    Arabic books & 495  \\ \hline
    Unique foreign terms &  58,570  \\ \hline
    All occurrences of foreign terms & 84,242 \\ \hline
    Total candidates & 334,564 \\ \hline
    Average number of candidates per occurrence & 3.97 \\ \hline 
    \end{tabular}

\end{table}

\newcolumntype{C}{>{\centering\arraybackslash}m{3.5cm}}
\begin{table}[t]
    \caption{Statistics of MASRAD (annotated candidates)}
    \label{t:masrad}
    \centering
    \small
\vspace{-1em}
    \begin{tabular}{|C|C|}
    \hline
    Arabic books & 15  \\ \hline
    Unique foreign terms &  4,347  \\ \hline
    All occurrences of foreign terms & 4,841 \\ \hline
    Total candidates & 19,405 \\ \hline
    \end{tabular}

\end{table}

The source dataset is derived specifically from books 
published by a set of publishing houses that we keep anonymous 
for blind review. 
These books span multiple academic fields, offering a diverse range of terminologies. The dataset includes terms in foreign languages (usually enclosed in parentheses) alongside their surrounding Arabic text. Table~\ref{t:books} summarizes the data extracted from all books. Table~\ref{t:masrad} summarizes the data in MASRAD, our labeled and expert reviewed data that will be used later for training MASRAD-Ex.

The labeling process involved extracting candidate Arabic terms 
preceding each foreign term and annotating them with a binary 
label: \textit{True} if the candidate is the correct translation 
of the foreign term, and \textit{False} otherwise. 
To streamline this process, a draft of labels was generated using 
heuristic-based rules, and were subsequently reviewed by experts
to  correct.

\subsection{Source data selection}

To ensure diversity and relevance, we selected texts from multiple domains. 
The candidate terms were extracted by identifying all strings of words, with a certain relative length with respect to the foreign term, immediately preceding a parenthetical foreign term. This ensures that potential translations are almost always captured for further analysis. 

\subsection{Annotated dataset construction}
Each extracted candidate term, along with its associated foreign term, was presented as an individual instance. 
The features of this instance, extracted using NLP tools, are described in \autoref{m}.

Subsequently, a draft of the labels was generated using the heuristics detailed in \autoref{h} to facilitate the labeling process. 
This draft was then reviewed to correct errors.

\section{Methodology}
\label{m}

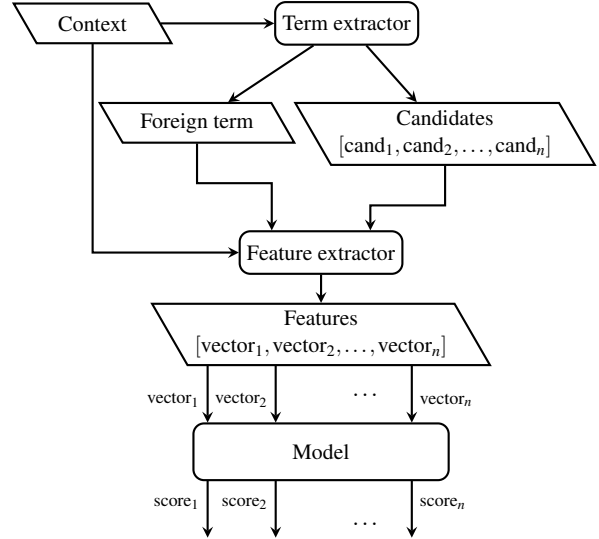
\begin{figure}[t]
    \centering
    \begin{tikzpicture}[node distance=1cm, scale=0.75, every node/.style={transform shape}]
    \tikzstyle{action} = [rectangle, rounded corners, minimum width=2cm, minimum height=.75cm, text centered, draw=black, thick]
    \tikzstyle{wideaction} = [rectangle, rounded corners, minimum width=4.5cm, minimum height=1cm, text centered, draw=black, thick]
    \tikzstyle{result} = [trapezium, trapezium left angle=120, trapezium right angle=60, minimum width=1.8cm, minimum height=0.7cm, text centered, draw=black, thick]
    \tikzstyle{arrow} = [thick,->,>=stealth]
    
    \node (context) [result] {Context};
    \node (extractor) [action, right=2cm of context] {Term extractor};
    \node (foreignterm) [result, below left=1cm and 1cm of extractor] {Foreign term};
    \node (candidates) [result, below right=1cm and -.1cm of extractor, align=center] {Candidates \\ $[\text{cand}_1, \text{cand}_2, \dots, \text{cand}_n]$};
    
    \node (features) [action, below=1.8cm of $(foreignterm)!0.5!(candidates)$] {Feature extractor};
    \node (featureoutput) [result, below=.5cm of features, align=center] {Features \\ $[\text{vector}_1, \text{vector}_2, \dots, \text{vector}_n]$};
    
    \node (model) [wideaction, below=1cm of featureoutput] {Model};
    
    
    
    \draw [arrow] (context.east) -- (extractor.west);
    \draw [arrow] (extractor) -- (foreignterm);
    \draw [arrow] (extractor) -- (candidates);
    
    \draw [arrow] (foreignterm.south) |- ($(foreignterm.south) + (0, -0.7)$) -| ($(features.north west)!0.2!(features.north east)$);
    \draw [arrow] (candidates.south) |- ($(candidates.south) + (0, -0.7)$) -| ($(features.north west)!0.8!(features.north east)$);
    
    \coordinate (context_left) at ($(features.west) - (1.2cm, 0)$);
    \draw [arrow] (context.south) |- (context_left) -- (features.west);
    
    \coordinate (candidates_mid) at ($(candidates.south) - (0, 3.5cm)$);
    
    \draw [arrow] (features) -- (featureoutput);
    
    \def\xshift{-2cm}
    \def\dx{1.2cm}
    
    \draw [arrow] ($(featureoutput.south)+(\xshift,0)$) -- node[pos=0.6, left] {\small $\text{vector}_1$} ($(model.north)+(\xshift,0)$);
    \draw [arrow] ($(featureoutput.south)+(\xshift+\dx,0)$) -- node[pos=0.6, left] {\small $\text{vector}_2$} ($(model.north)+(\xshift+\dx,0)$);
    
    \node at ($(featureoutput.south)!0.5!(model.north)+(\xshift+2*\dx+0.4cm,0)$) {\large $\dots$};
    
    \draw [arrow] ($(featureoutput.south)+(\xshift+3*\dx,0)$) -- node[pos=0.6, right] {\small $\text{vector}_n$} ($(model.north)+(\xshift+3*\dx,0)$);
    
    

    \def\arrowlength{1cm} 
    
    \draw [arrow] ($(model.south)+(\xshift,0)$) -- ++(0,-\arrowlength) node[pos=0.4, left] {\small $\text{score}_1$};
    \draw [arrow] ($(model.south)+(\xshift+\dx,0)$) -- ++(0,-\arrowlength) node[pos=0.4, left] {\small $\text{score}_2$};
    
    \node at ($(model.south)+(\xshift+2*\dx+0.4cm,-0.75*\arrowlength)$) {\large $\dots$};
    \draw [arrow] ($(model.south)+(\xshift+3*\dx,0)$) -- ++(0,-\arrowlength) node[pos=0.4, right] {\small $\text{score}_n$};

\end{tikzpicture}
\vspace{-2em}
    \caption{An overview of the process}
    \label{fig:process_overview}
\end{figure}

Figure~\ref{fig:process_overview} illustrates the flow of our methodology. 
First, the tool extracts the candidates from a context (usually a paragraph) with a foreign term. 
Then, the feature vector of each candidate is computed. Finally, each vector is fed to a model which outputs a score that helps in choosing the target translation of the source foreign term.

\subsection{Feature extraction}
Given a candidate $s_i$, its foreign term $f$, and its context $c$, we extract features of the candidate that are going to be used later to predict which candidate is the target term. 
These features with their possible values are listed in Table~\ref{t:feat} and explained in the following subsections.

\newcolumntype{C}{>{\centering\arraybackslash}m{3.2cm}}
\begin{table}[t]
    \centering
    \small
    \caption{Features extracted for each candidate
    \label{t:feat}}
    \vspace{-1em}
    \begin{tabular}{|C|C|}
    \hline
    \textbf{Feature} & \textbf{Value}  \\ \hline
    Semantic similarity & [0, 1]   \\ \hline
    Translation lexical &  \\ similarity & [0, 1]  \\ \hline
    Transliteration lexical & \\ similarity & [0, 1]  \\ \hline
    Entity & \{PER, LOC, ORG, MISC\} \\ \hline
    POS& \{adj, adv, conj, misc, noun, noun\_prop, part, prep, pron, verb\} \\ \hline
    Phonetic similarity & \{False, True\} \\ \hline
    \end{tabular}

\end{table}

\subsubsection{Semantic similarity}
\label{semantic}

We utilized the sentence transformer \cite{reimers-2019-sentence-bert} LaBSE \cite{feng2022languageagnostic} to compute the embeddings of the source term and candidate term. 

After that, we compute semantic similarity as the cosine similarity between these 2 vectors \cite{salton1975vector}. This might seem sufficient but it's not, which is due to the high similarity between candidates (given any two candidates, one of them is a substring of the other). Therefore, we need more features to handle peculiar cases.

\newcolumntype{C}[1]{>{\centering\arraybackslash}m{#1}}

\begin{table}[tb]
    \centering
    \caption{Semantic similarity example} 
    \label{t:anne}
    \vspace{-1em}
    \small
    \begin{tabular}{|C{4cm}|C{2cm}|}
    \hline
    \textbf{Candidate} & \textbf{Semantic similarity}  \\ \hline
    \textAR{آن هيرزبرغ} \break \textit{Gloss: "Anne Herzberg"} &  0.635 \\ \hline
    \textAR{هيرزبرغ} \break \textit{Gloss: "Herzberg"} &  0.66  \\ \hline
    \end{tabular}

\end{table}

Table~\ref{t:anne} shows how the semantic similarity alone score favors the $2^{nd}$ candidate, which missed the first name in the foreign term and is incorrect according to experts.

\subsubsection{Translation lexical similarity}
Translation lexical similarity uses automated translation
of the foreign term and measures a lexical distance between 
that and the candidate terms. 
The distance of choice is the Levenshtein~\cite{levenshtein1966} distance which computes
how many changes are needed for the terms to be identical. 
We use a ratio with respect to length to normalize the distance across candidates. 

Example: suppose the foreign term is ``London School of Economics and Political Science". 
Automatic translation returns  \\ $t=$\textAR{كلية لندن للاقتصاد والعلوم السياسية} \\
\textit{Gloss: "London School of Economics and Political Science."}

We have the candidate: \\ $s=$\textAR{كلية لندن للعلوم الاقتصادية والسياسية}  \\
\textit{Gloss: "London School of Economics and Political Science."} \\
The Levenshtein distance between $s$ and $t$ is 15. The ratio is calculated as 

$$\frac{\text{len}(t)+\text{len}(s)-15}{\text{len}(t)+\text{len}(s)} = 0.79$$

\subsubsection{Transliteration lexical similarity}
This feature captures the cases where the target Arabic term is just a transliteration of the source term by computing the Levenshtein ratio between the transliteration of the source term and the candidate.
This is helpful in  the case of translating proper nouns that should supposedly map phonetically.

Example: suppose the foreign term is "Ehud Prawer". After transliterating it, we get \\ $t=$\textAR{ايهود براور} \\
\textit{Gloss: "Ehud Brower"} \\
We also have the candidate with a hamza instead of an alef 
\\ $s=$\textAR{إيهود براور} \\
\textit{Gloss: "Ehud Brower"} \\
The Levenshtein distance between $s$ and $t$ is 1. The ratio is  calculated as $$\frac{\text{len}(t)+\text{len}(s)-1}{\text{len}(t)+\text{len}(s)} = 0.95$$

\subsubsection{Entity}
A named entity is a real-world object or concept that can be clearly identified and categorized with a proper name.

Candidates that are detected as named  entities  have a higher probability of being a term. 
We use an open source Arabic  NER model ~\cite{jarrar2022wojood} to detect entities in Arabic context. 
Also, NER is used on the source term to favor candidates having the same entity type; we used the spaCy ~\cite{spacy2020} model xx\_ent\_wiki\_sm in particular.

\subsubsection{Part of speech}
A part of speech (POS) is a category of words with similar grammatical properties. Common parts of speech include nouns, verbs, adjectives, adverbs, pronouns, prepositions, conjunctions, and interjections.
Terms are mostly noun phrases and rarely start with verbs. 

We used POS tagging tools~\cite{obeid-etal-2020-camel} with its maximum likelihood expectation morphological disambiguator 
to detect the POS of the first word occurring in a candidate. 

An alternative is to use the POS tagging tools by~\cite{MagedSaeed:farasapy}.

\subsubsection{Phonetic similarity}

As illustrated in Table~\ref{t:phonetic}, we use the soundex algorithm~\cite{russell1918soundex} to determine whether the source and candidate terms sound the same. 
This serves a  similar purpose to  transliteration lexical similarity; however, practicaly characterized  with  higher precision and lower recall.

\newcolumntype{C}{>{\centering\arraybackslash}m{2.5cm}} 


\begin{table}[tb]
    \caption{Phonetic similarity examples}
    \label{t:phonetic}
    \centering
    \small
    \resizebox{.49\textwidth}{!}{
    \renewcommand{\arraystretch}{1.3} 

    \begin{tabular}{|C|C|C|}
    \hline
    \textbf{Source term} & \textbf{Candidate} & \textbf{Phonetically similar} \\ \hline 
    Regavim & \textAR{ريغافيم} \break \textit{Gloss: "Regavim"} & True  \\ \hline
    Lockheed Martin &  \textAR{لوكهيد مارتن}  \break \textit{Gloss: "Lockheed Regavim"} & True \\ \hline
    Lockheed Martin & \textAR{مثل لوكهيد مارتن} \break \textit{Gloss: "like Lockheed Regavim"} & False \\ \hline
    \end{tabular}
}    
\end{table}

\section{Results}
\label{s:results}

\subsection{Heuristics}
\label{h}
In this approach, we came up with the following formula after multiple iterations of observing the results and fine-tuning:

\[
s = S_L + S_S + S_E + S_P + S_{POS}
\]

1. \textbf{Lexical score} \( S_L \):
\[
S_L = \mathbb{1} \left( l_1 \geq \tau \right) \cdot (1.2 \cdot l_1) + \mathbb{1} \left( l_2 \geq \tau \right) \cdot l_2
\]
where \( l_1 \) is the lexical similarity with the translation, \( l_2 \) is with the transliteration, and \( \tau = 0.7 \) is the similarity threshold.

2. \textbf{Semantic score} \( S_S \):
\[
S_S = 1.45 \cdot S
\]
where \( S \) is the semantic similarity score.

3. \textbf{Entity score} \( S_E \):
\[
S_E =
\begin{cases}
0.7, & e \neq \text{"none"} \text{ and } e = e_s \\
0.3, & e \neq \text{"none"} \text{ and } e \neq e_s \\
0, & \text{otherwise}
\end{cases}
\]
where \( e \) is the candidate's entity label and \( e_s \) is the source term's entity label.

4. \textbf{Phonetic score} \( S_P \):
\[
S_P =
\begin{cases}
1, & \text{if phonetically similar} \\
0, & \text{otherwise}
\end{cases}
\]

5. \textbf{Part of speech score} \( S_{POS} \):
\[
S_{POS} =
\begin{cases}
-1, & \text{if pos} \in \{ \text{"verb"}, \text{"prep"}, \text{"conj"} \} \\
1, & \text{if pos} \in \{ \text{"noun"}, \text{"noun\_prop"} \} \\
0, & \text{otherwise}
\end{cases}
\]

\vspace{0.3em}
\( \mathbb{1}(A) \) is the indicator function:
\[
\mathbb{1}(A) =
\begin{cases}
1, & \text{if } A \text{ is true} \\
0, & \text{otherwise}
\end{cases}
\]

After that, we select the candidate with the maximum score within an occurrence of a foreign term. 
Table~\ref{t:unified_heuristics_ml_results} shows the  results achieved by the heuristics approach on the whole dataset. Note that fine-tuning was done only by observing the training data. These decent results provided a good starting point that later helped in the labeling process of the data.


\subsection{Machine learning}

An instance of the dataset is the features of a candidate,
and a label that is \textit{True} if the candidate is the target candidate and \textit{False} otherwise.

We also augmented the feature space with the following: 
semantic similarity rank within the candidates related to the same {\em occurrence} of a foreign term, semantic similarity difference with the candidate ranked $1^{st}$, translation lexical similarity rank, translation lexical similarity difference, transliteration lexical similarity rank, transliteration lexical similarity difference.
The intuition behind the augmentation is to provide comparative context to the ML model with respect to other candidates. 

Let $x \in$ \{semantic similarity, translation lexical similarity, transliteration lexical similarity\}, \\
$A_{i}^{x}$ = Array of values of feature $x$ of candidates related to the same {\em occurrence} $i$ of a term, sorted by decreasing order. \\ 
For each candidate $j$:
\[
\text{x\_rank}(j) = A_{\text{occ}(j)}^{x}.\text{index\_of}(x(j))
\]
\[
\text{x\_difference}(j) = \max(A_{\text{occ}(j)}^{x}) - x(j)
\]

where the candidate j belongs to the {\em occurrence} occ(j).

As shown in Table~\ref{t:books}, We annotated 19,405 candidates to build the dataset. 
Then we trained a binary classifier  using the Auto-WEKA package on the Weka Workbench~\cite{frank2016weka}. 
We allowed Weka to run for several hours on 8 threads, which resulted in the arguments displayed in Table ~\ref{tab:weka}.

\begin{table}[htbp]
  \centering
  \renewcommand{\arraystretch}{1.5}
  \small
  \caption{AutoML arguments}
  \label{tab:weka}
  \begin{tabularx}{\columnwidth}{@{}lX@{}}
    \textbf{Best classifier:}               & \texttt{RandomForest} \\
    \textbf{Arguments:}                     & \texttt{[-I, 252, -K, 0, -depth, 0]}         \\
    \textbf{Attribute search:}              & \texttt{GreedyStepwise} \\
    \textbf{Attribute search arguments:}    & \texttt{[-C, -B, -R]}                            \\
    \textbf{Attribute evaluation:}          & \texttt{CfsSubsetEval} \\
    \textbf{Attribute evaluation arguments:}& \texttt{[-L]}                                \\
    \textbf{Metric:}                        & \texttt{errorRate}                           \\
  \end{tabularx}
\end{table}

\noindent
Table \ref{t:unified_heuristics_ml_results} summarizes the results. The training set comprises data from nine books, and the test set comprises data from six books. We split the data at the book level to make the task more challenging, where the model is tested on books it has never seen before, which better reflects a real-world scenario.

We also tried ranking candidates by partitioning the testing dataset into groups of candidates related to the same occurrence of a term, and ranking them based on the value of their prediction, and then the candidate with the highest rank was chosen as the target term, and the others where detected as False. This achieved a 90.1\% precision and 92.9\% recall.


\begin{table}[tb]
\caption{Comparison of Heuristics and ML Results}
\label{t:unified_heuristics_ml_results}
\vspace{-1em}
\centering
\resizebox{\columnwidth}{!}{%
\begin{tabular}{|c|c|c|c|}
\hline
\textbf{Category}      & \textbf{Metric} & \textbf{Training} & \textbf{Testing} \\ \hline
\multirow{3}{*}{Heuristics} 
                           & Precision      & 88.8\%      & 81.5\% \\ \cline{2-4} 
                           & Recall         & 88.2\%      & 84.4\% \\ \cline{2-4} 
                           & F1 Score       & 88.5\%      & 82.9\% \\ \hline 
\multirow{3}{*}{ML}        
                           & Precision      & 100\%      & 90.5\% \\ \cline{2-4} 
                           & Recall         & 100\%      & 92.4\% \\ \cline{2-4} 
                           & F1 Score       & 100\%      & 91.4\% \\ \hline
\multicolumn{2}{|c|}{\textbf{Dataset Size}} & 15865 & 3540 \\ \hline
\end{tabular}%
}

\end{table}

\subsection{Discussion and error analysis}

Upon further inspection, it was observed that most of the classification errors made by the ML model involved cases that required expert judgment to discern their correctness. This highlights an advantage of the model: its errors occur primarily on inherently difficult cases, which are challenging even for experts. Table~\ref{t:hard_error} shows an example.
\newcolumntype{C}{>{\centering\arraybackslash}m{3.5cm}}
\newcolumntype{c}{>{\centering\arraybackslash}m{2cm}}
\begin{table}[tb]
\caption{Selecting a wrong candidate for the foreign term \textit{SA} (\textit{Gloss: "Sturmabteilung"})}
\label{t:hard_error}
\vspace{-1em}
\centering
\resizebox{.49\textwidth}{!}{
\begin{tabular}{|C|c|c|}
\hline
\textbf{Candidate} & \textbf{Prediction} & \textbf{Label} \\
\hline
\textAR{كتيبة العاصفة}\break\textit{Gloss: "Storm Battalion"} & False & True \\
\hline
\end{tabular}
}
\end{table}

Another cause of errors are peculiar cases where there is one letter difference between the correct translation and the selected candidate, which is caused by Arabic prepositions linked to a word. This is usually handled through pre-processing or providing both candidates to the model for it to decide on one, which might fail in some cases. Table~\ref{t:prep_error} shows an example.

\newcolumntype{C}{>{\centering\arraybackslash}m{3.5cm}}
\newcolumntype{c}{>{\centering\arraybackslash}m{2cm}}
\begin{table}[tb]
\centering
\caption{Selecting a wrong candidate for the foreign term \textit{Verteilungsstellen} (\textit{Gloss: "distribution points"})}
\label{t:prep_error}
\vspace{-1em}
\resizebox{.49\textwidth}{!}{
\begin{tabular}{|C|c|c|}
\hline
\textbf{Candidate} & \textbf{Prediction} & \textbf{Label} \\
\hline
\textAR{هيئات التوزيع}\break\textit{Gloss: "distribution entities"} & True & True \\
\hline
\textAR{بهيئات التوزيع}\break\textit{Gloss: "by distribution entities"} & True & False \\
\hline
\end{tabular}
}
\end{table}

\subsection{Evaluation against a manual glossary}
\label{glossary}

Over 495 books, our extraction pipeline identified 58,570 unique foreign terms.
Intersecting these with our manual in‐house glossary of 46,787 concepts (a concept here consists of an English, a French, and an Arabic term) yielded 7,368 matches (15.74\% of the glossary). This low overlap highlights a major problem in terminology management: assuming in the worst case that a concept was matched with both the English and French foreign terms of our extracted data, we get that at least 43,834 (74.84\%) of the extracted terms are novel and need to be handled.

Focusing on the 7,368 {\em matched concepts}, we evaluated our ranking model’s ability to surface the
expert Arabic term. The correct glossary entry appeared as the top automated suggestion 30.6\% of the time (top-1 accuracy), within the top two automated suggestions 32.99\% of the time (top-2 accuracy), and within the top three automated suggestions 33.3\% of the time (top-3 accuracy).

By observing these accuracies, we can assume that the percentage of manual Arabic terms in these {\em matched concepts} that are also present in the 495 processed books to be close to our top-3 accuracy (33.3\%). This reasoning is due to the minimal jumps between the top-k accuracies. From this we deduce two key insights: 
\begin{itemize}
    \item About 66.7\% of the {\em matched concepts} have a different Arabic translation by the authors or translators of the processed books than that of the expert. This greatly highlights the inconsistency problem in terminology.
    \item Among the {\em matched concepts} with a present Arabic term in the books, our model placed it first in 91.89\% of the cases (top-1 accuracy ÷ top-3 accuracy). This indicates that our model effectively elevates the expert translation when it is available. Note that the model performs really well when applying it to problems since it aggregates and compares results from the candidates of all occurrences of a term.
\end{itemize}

\section{Related work}
\label{s:rel}
TURJUMAN~\cite{nagoudi2022turjuman} is a neural machine translation toolkit that translates into Modern Standard Arabic using fine-tuned AraT5 models. While effective at sentence-level translation, it does not address terminology alignment or consistency in specialized texts.

In addition to bilingual dictionaries, efforts have been made to develop lexical sources from several original languages into Arabic.
Arabic WordNet~\cite{arabic_wordnet} for instance relies on several open and restricted domain sources to develop a lexical English-Arabic database in different fields.
Other more specialized glossaries have been developed, such as Hossam Mahdi's Glossary of Terms for the Conservation of Cultural Heritage in Arabic Alphabetical Order~\cite{iccrom_glossary}.
Furthermore, glossaries offering support to interpreters \cite{awara85_blog} have also been established. 

The Arabic digital ontology~\cite{arab-ontology-jarrar-2022} includes digitized Arabic dictionaries as denoted by  dictionary authors. 
It requires API access for term lookup. 

However, these are not complete, 
need to be updated manually regularly for novel terms, and are not easy to integrate in the editing process.

\section{Conclusion}

This paper introduced the semi-automatic construction of MASRAD using MASRAD-Ex, a set of tools for extracting and aligning terminology. 
Our work achieves useful results, contributing to automated \emph{terminology management} and text processing. 
The created dataset and corpora further enrich the research community, advancing cross-lingual knowledge exchange. 

\section*{Limitations and future work}
Some terms have subtle translations that are inherently complex and challenging to identify, even for human experts. Therefore, the tool is designed to assist rather than replace expert judgment, effectively handling simpler terms automatically while supporting experts with more difficult cases.

As shown in Section~\ref{glossary}, 84.26\% of the entries in the manual glossary were not matched. This highlights the richness of the world of terminology and the necessity of drawing on additional sources to expand our termbase. To this end, we plan to apply MASRAD-Ex to newly acquired books and expand our tooling to support other type of sources.

To improve the semantic similarity measure~\ref{semantic}, we can fine-tune a sentence transformer \cite{reimers-2019-sentence-bert}, or train a small model that augments the space of an existing one.

\section*{Ethics statement}
The data was collected and used with the 
appropriate approvals of the intellectual 
property owners. 
All results are reported following 
best academic standards and practices. 

\bibliography{refs}

\end{document}